\documentclass[fleqn,10pt]{article}
\usepackage{microtype}
\usepackage{setspace}
\usepackage{hyperref} 
\usepackage{arxiv}
\usepackage[utf8]{inputenc} 
\usepackage[T1]{fontenc}    
\usepackage{url}            
\usepackage{booktabs}       
\usepackage{amsfonts}       
\usepackage{nicefrac}       
\usepackage{microtype}      
\usepackage{lipsum}         
\usepackage{graphicx}
\usepackage{cite}
\usepackage{doi}
\usepackage{amsmath,amssymb,amsfonts}
\usepackage{algorithmic}
\usepackage{graphicx}
\usepackage{textcomp}
\usepackage{xcolor}
\def\BibTeX{{\rm B\kern-.05em{\sc i\kern-.025em b}\kern-.08emT\kern-.1667em\lower.7ex\hbox{E}\kern-.125emX}}

\usepackage{booktabs} 
\usepackage{graphicx}
\usepackage{epstopdf}
\usepackage{url}
\usepackage{algorithmic}
\usepackage{multirow}
\usepackage{caption}
\usepackage{amssymb}
\usepackage{amsfonts}
\usepackage{float}
\usepackage{bm}
\usepackage{dsfont}
\usepackage[T1]{fontenc}
\usepackage[left,modulo]{lineno}
\usepackage[ruled, vlined, linesnumbered, longend]{algorithm2e}
\usepackage{psfrag}
\usepackage{subfigure}
\usepackage{fancyhdr}
\usepackage{eucal}
\usepackage[english]{babel}
\usepackage{ifpdf}
\usepackage{cleveref} 
\usepackage{textcomp}
\usepackage{tikz}
\usepackage{tikz-inet}
\usetikzlibrary{calc,shapes.geometric}
\usetikzlibrary{shapes,snakes,arrows,automata} 
\usetikzlibrary{patterns}
\usepackage{multirow}
\usepackage{multicol}
\usepackage{authblk}
\newcommand{\by}{\mathbf{y}}

\newcommand{\yNt}{\overline{\by}_{w90,t}}
\newcommand{\ynt}{\overline{y}_{w90,t}}
\newcommand{\yns}{\overline{y}_{w90,s}}

\newcommand{\TP}{\mathit{TP}}
\newcommand{\FP}{\mathit{FP}}
\newcommand{\TN}{\mathit{TN}}
\newcommand{\FN}{\mathit{FN}}

\begin{document}
%

%
\title{An AutoML Framework using AutoGluonTS\\
for Forecasting Seasonal Extreme Temperatures
}


%

\newbox{\orcid}\sbox{\orcid}
{\includegraphics[scale=0.06]{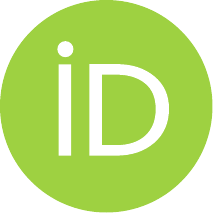}} 

%

\author{
\parbox{\textwidth}{
  \href{https://orcid.org/0000-0001-5527-5798}{\includegraphics[scale=0.08]{orcid.pdf}}\hspace{1mm}Pablo Rodr\'iguez-Bocca\textsuperscript{1}, Guillermo Pereira\textsuperscript{1}, 
  Diego Kiedanski\textsuperscript{1}, 
   \href{https://orcid.org/0000-0001-7785-2998}{ \includegraphics[scale=0.08]{orcid.pdf}}Soledad Collazo\textsuperscript{2,3},
   \href{https://orcid.org/0000-0002-9172-0155}
  {\includegraphics[scale=0.08]{orcid.pdf}}Sebasti\'an Basterrech\textsuperscript{4}, 
  \href{https://orcid.org/0000-0002-1712-0477}{\includegraphics[scale=0.08]{orcid.pdf}}Gerardo Rubino\textsuperscript{5}\\
  \textsuperscript{1}Facultad de Ingenier\'{\i}a, Universidad de la Rep\'ublica, Montevideo, Uruguay\\
  \textsuperscript{2}Departamento de Física de la Tierra y Astrofísica, Universidad Complutense de Madrid, Spain\\
  \textsuperscript{3}Faculty of Exact and Natural Sciences, Universidad de Buenos Aires (UBA), Argentina\\
  \textsuperscript{4}Technical University of Denmark, Kongens Lyngby, Denmark\\
  \textsuperscript{5}INRIA, Rennes, France\\
  Emails: \texttt{prbocca@fing.edu.uy}, \texttt{guillermopereira495@gmail.com}, \texttt{dkiedanski@gmail.com},\\
  \texttt{scollazo@ucm.es}, \texttt{sebbas@dtu.dk}, \texttt{gerardo.rubino@inria.fr}
}}
\date{}
\renewcommand{\undertitle}{To appear in the proceedings of IJCNN 2025\\ Workshop: AI for a Cooler Planet: Tackling Environmental Challenges with Neural Networks}
\renewcommand{\shorttitle}{An AutoML Framework for Forecasting Seasonal Extreme Temperatures}

\maketitle

\begin{abstract}
In recent years, great progress has been made in the field of forecasting meteorological variables.
Recently, deep learning architectures have made a major breakthrough in forecasting the daily average temperature over a ten-day horizon.
However, advances in forecasting events related to the maximum temperature over short horizons remain a challenge for the community.
A problem that is even more complex consists in making predictions of the maximum daily temperatures in the short, medium, and long term.
In this work, we focus on forecasting events related to the maximum daily temperature over medium-term periods (90 days). Therefore, instead of addressing the problem from a meteorological point of view, this article tackles it from a climatological point of view.
Due to the complexity of this problem, a common approach is to frame the study as a temporal classification problem with the classes: maximum temperature \textit{above normal}, \textit{normal} or \textit{below normal}.
From a practical point of view, we created a large historical dataset (from 1981 to 2018) collecting information from weather stations located in South America. 
In addition, we also integrated exogenous information from the Pacific, Atlantic, and Indian Ocean basins.
We applied the AutoGluonTS platform to solve the above-mentioned problem. 
This AutoML tool shows competitive forecasting performance with respect to large operational platforms dedicated to tackling this climatological problem; but with a ``relatively'' low computational cost in terms of time and resources.
%
%
%
%
%
%
\end{abstract}


\section{Introduction}\label{s:intro}

Forecasting extreme seasonal temperature events is especially relevant for society, for example, in agriculture, as well as other critical areas such as public health care~\cite{troccoli2008}.
Improving its accuracy on the seasonal scale would have a huge impact on our lives.
Today, most of the techniques proposed in the literature correspond to process-based dynamic models and classical statistical methods.
However, innovative approaches are emerging from the Artificial Intelligence area that are promising in tackling this kind of forecasting problem. 
Recently, two Deep Learning architectures achieved outstanding performance, when we compare them to the state-of-the-art methods for the meteorological forecasting issue of predicting the weather in the following ten days~\cite{Pangu-weather23,graphcast23}.
Forecasting extreme temperatures in the future is an important problem, well-known for its modelling challenges~\cite{Collazo2019}.
There have been significant advances in forecasting daily average temperatures. However, progress in forecasting extreme temperatures has been more limited during the last years.
A common methodological approach consists of simplifying the problem. Instead of forecasting precise maximum daily temperature over medium-term periods, the task is to predict a category level of the maximum daily temperature during that period.
Therefore, the task is transformed to a multi-class classification problem with spatio-temporal data. The commonly used classes define three levels: maximum temperature \textit{above normal}, \textit{normal} or \textit{below normal}.
In this paper, we address this challenge of predicting the maximum temperature levels in a medium-term horizon of 90 days.
The study case covers a large area in the southern area of South America, including regions in Argentina, Chile, Brazil, Paraguay, and Uruguay. 
There are two main sources of information for the investigated historical data that covers 40~years. 
We collect meteorological data from regional weather stations spread accross the area (137 weather stations). 
The stations were chosen to cover a diverse range of climates.
In addition, we also integrated exogenous variables, including information on sea surface temperature from Pacific, Atlantic and Indian Ocean basins, which may affect climate variability and weather patterns in South America~\cite{Collazo2019}.

After a long process of initial evaluation of different learning methods, we chose to use AutoGluon-Time Series (AutoGluonTS)~\cite{AG-Tab20}, which was already successfully used for forecasting \textit{average temperatures} in these regions~\cite{Kiedanski2025}.
AutoGluonTS, a framework from the automated machine learning (AutoML) category, fits various data-driven models ranging from ARIMA and decision trees to neural network models.
AutoGluonTS offers a powerful alternative to other deep learning tools for spatio-temporal modeling, such as XGBoost and Graph NNs, as it automates important aspects of the architecture setting and configuration process.

In summary, the main contributions and novelties of this article are two-fold: (i)~the creation of a high-quality, large historical dataset from different regional climates and sufficient information for studying extreme temperature patterns, and (ii)~an empirical study of the potential of the recently introduced AutoGluonTS tool for tackling the problem of predicting extreme temperatures.
These contributions are particularly significant to the climate informatics community because there is a scarcity of studies focusing on South America, unlike the numerous investigations conducted in the Northern Hemisphere~\cite{Collazo2019}.

The rest of this article is organized as follows.
Section~\ref{s:background} presents the specification of the problem of forecasting extreme temperatures and a review of related work. Material and methods are presented in Section~\ref{sec:matandmet}. Result analysis is discussed in Section~\ref{sec:results}. Finally, Section~\ref{sec:conclusions} provides concluding remarks and future research directions.

\section{Background}\label{s:background}

\subsection{Specification of the problem}

%

In this work, we have data coming from a large set of stations that provide daily values for several variables (that is, not only temperature, see below).
We start by specifying the notation of the day-based sequential encoding used in our study.
The unit of time being the day, we have, in particular, the maximum temperature observed in day~$t$, which we denote by~TX$_t$. 
The collected data starts in Jan~1~1981, which is day~0; Jan~2~1981 is day~1, Dec~31~1981 is day~364, Jan~1~1982 is day~365, etc. 
Given a day number $t \in \mathbb{N}$, we can define year($t$) as the corresponding year.
For instance, year(7) = 1981, year(400) = 1982, etc. Next, strings such as ``Jan~1'' or ``Mar~27'' are \textit{calendar dates}, and we define cal($t$) as the calendar date corresponding to day~$t$; examples: cal(10) = Jan~11, cal(11) = Jan~12, cal(375) = Jan~11, etc. 
Finally, we call \textit{complete dates} strings such as ``Jan~1~1981'' or ``Apr~7~2002''. We associate with this definition function date($t$) as the concatenation of cal($t$), ``\space'' and year($t$). So, date(10) = Jan~11~1981 and date(375) = Jan~11~1982.
%
%

A day $t$ is defined as a \textit{warm day} when the maximum temperature of this day is above the~1981–2010 90th percentile. The specialised literature considers the range of years from 1981 to 2010 adequate to capture the distribution of climatic data~\cite{Sulikowska2020}; for this reason, we also use it as a reference in this work. 
For that, we consider the maximum temperature of the~30 days~$s$, in our dataset, such that~cal($s$) = cal($t$) belonging to the reference period going from Jan 1 1981 to Dec 31 2010. To improve the learning process, we add to these values the maximum temperature of the two days before cal($s$) and the two days after cal($s$). Denote by D($t$) this set of 150~values, we compute the 90th percentile value over this, and define the variable $y_t \in \{0,1\}$ if the day $t$ is warm or not, denoted as TX90 (common notation in the area, e.g.,\cite{Sulikowska2020}).

At each time $t$, we define the percentage of warm days in the last~$90$ days window, denoted as TX90w90, in a specific geographical point, as
\begin{equation}
    \ynt = \frac{1}{90} \sum_{i=t-89}^t y_i. 
\end{equation}

It is important to note the following: if cal($t$) =  Feb 28 then TX90w90 represents the seasonal percentage of warm days in December–February, a typical measure in the literature, denoted as DJF TX90, that follows the recommendations of the WMO Joint Commission for Climatology (CCL) /CLIVAR/JCOMM Expert Team on Climate Change Detection and Indices (ETCCDI)~\cite{ETCCDI}.

In current state-of-the-art, the forecasting of TX90w90 values is a very ambitious task. Instead, the task performed by the specialized literature is to predict a category level of this percentage of warm days.
We must classify the percentage of warm days until day~$t$ as being in one of the three classes \textit{below normal}, \textit{normal} and \textit{above normal}, regarding~$\ynt$. 
For that, we follow the same procedure as above, 
we consider the percentage of warm days, $\yns$, in the set E($t$) of 150 days $s$ in the reference dataset that goes from Jan 1 1981 to Dec 31 2010, such that~$\text{cal}(t) \in \{\text{cal}(s+i), i \in [-2,\ldots,2] \}$.
Over E($t$), we compute the first and second terciles as $\tau_{1,t}$ and $\tau_{2,t}$, respectively. Then, day~$t$ is \textit{below normal} if $\ynt \leq \tau(1,t)$, it is \textit{normal} if $\tau(1,t) < \ynt \leq \tau(2,t)$, and \textit{above normal} if $\tau(2,t) < \ynt$. We can define the function $\text{class}(\ynt) \in \{\text{below normal}, \text{normal},\text{above normal} \}$ to compute these classes.

We have a set~$\mathcal{P}$ of~$N = 137$ geographical points with weather stations, where we want to predict events associated with the temperature in the following~$90$ days.
At each geographical point, we have historical data consisting of daily measures of maximum and minimum temperature (in Celsius), accumulated precipitations (in mm), etc.
The categorization of the percentage of maximal temperatures is done for all geographical points and calendar days. 
Therefore, we have a class level of the TX90w90 value for each day, $t$, and each geographical point, denoted as $\yNt \in \mathbb{R}^{N}$.

As TX90w90 is an average of values in the last 90 days,
predicting at time~$t$ the percentage of warm days in the following 90 days is the same as predicting the TX90w90 for time $t+90$.
Considering this, we can formulate our forecasting problem.
In the training phase, the data consist of the vectors of the percentage of warm days $\yNt$ and exogenous variables denoted by $\mathbf{X}_{t} \in \mathbb{R}^{N \times F}$, where $F$ denotes the number of exogenous variables.
Exogenous variables include accumulated precipitations, atmospheric pressure, and so on, which are described in~\ref{sec:dataset}.
The predictive problem consists in finding a parametric function $f(\cdot)$ that predicts the target class $S$ days ahead, based on historical data from the last $T$ days
\begin{equation}
\mathbf{y}{(t-T):t}, \mathbf{X}{(t-T):t} \xrightarrow{f} \text{class}(\mathbf{y}_{t+S}), 
\end{equation}
where $\mathbf{y}{(t-T):t} \in \mathbb{R}^{N \times T}$, $\mathbf{X}{(t-T):t} \in \mathbb{R}^{N \times F \times T}$, and $\mathbf{y}_{t+S} \in \mathbb{R}^{N}$. 
In our study, we fix the time horizon in 90 days ($S = 90$), that is, we predict the class of percentage warm days in the following 90 days, and the size of the historical data used for training, $T$, depends on the model.

\subsection{Related work}
Regarding the seasonal forecast of extreme temperature events, a model for predicting the summer heat waves over the Eurasian continent based on machine learning (ML) was investigated in~\cite{Zhang_2022}. 
In~\cite{FISTER2023110118} several CNN-based models and decision tree-based models were used to investigate the problem of extreme air temperature forecasting in two really different regions: Paris, France, and Córdoba, Spain.
In recent years, several dedicated models have been developed for weather and climate forecasting. These models combine statistical methods, dynamical systems, physical models, and machine learning approaches. The research institutions that develop these large-scale forecasting models provide regular reports and forecasting results online. Here, we review some of the most important references.
The North American Multi-Model Ensemble (NMME) tool is a large predictive platform and a global reference~\cite{NMME}, which integrates simulated data, chaos modeling, statistics,  and large learning models. 
The Climate Predictability Tool (CPT), from Columbia University~\cite{SMN-CPT}, combines historical data from atmospheric information, wind, temperature, and other meteorological variables.
Foundation models have been used in the ClimaX tool. This large model showed flexibility, reliability, and generalizability properties for weather and climate predictions~\cite{CLIMAX}.
The area of AutoML has gained enormous relevance in recent years. Advances in computational power, high-performance machines, and optimization techniques have significantly enhanced the performance of ML models within reasonable computational times.
Among the most powerful public AutoML frameworks available is AutoGluonTS~\cite{Erickson2023}, which we used in our study.
For a more comprehensive overview of AutoML frameworks, including tools specialized in time series problems, we recommend the book~\cite{Hutter2019} and the survey articles~\cite{He2019, Yao2018, Truong2019, Zoller2019}.

\section{Materials and methods}\label{sec:matandmet}
\subsection{Evaluation  metrics}\label{sec:metrics}

The case of extreme temperature prediction can be seen as a three-class classification problem at any future time and at any station.
Then we analyze the results using standard classification metrics.
We use True Positives ($\TP$), False Positives ($\FP$), of True negatives ($\TN$) and False Negatives ($\FN$). 
In addition, we also evaluate the $\text{precision}$, $\text{recall}$ and $\text{F1-score}$ scores. 
Let $C$ be the set of classes $\{\text{above normal, normal, below normal}\}$. For each~$c$ in $C$, we have the scores~$\TP_c$, $\FP_c$, $\TN_c$, and $\FN_c$. 
We recall the following two global performance metrics:
\begin{equation}
 \text{F1-score}_{\text{macro}} = \frac{1}{|C|} \sum_{c \in C} \frac{2 TP_c}{2 \TP_c + \FN_c + \FP_c},
 \label{eq:f1macro}
\end{equation}
and
\begin{equation}
 \text{F1-score}_{\text{micro}} = \frac{\sum_{c \in C} 2 \TP_c}{ \sum_{c \in C} (2 \TP_c+ \FN_c + \FP_c) }.
 \label{eq:f1micro}
\end{equation}
Note that~(\ref{eq:f1macro}) considers the average of the F1-scores for each class, and~(\ref{eq:f1micro}) considers all the classes simultaneously. For this reason, they are referred to as \textit{macro-averaging} and \textit{micro-averaging} scores, respectively~\cite{opitz2021macrof1macrof1}.

Furthermore, we use the Relative Operating Characteristics (ROC) curve and its area under the curve (AUC), which are commonly used in classification problems. The ROC curve plots the TP rate against the FP rate when the forecast probability of the positive event varies. 
The AUC is the integral of the ROC curve. An area of~$1.0$ means a perfect predictor, and~$0.5$ corresponds to a predictor with no discriminative ability~\cite{mason2018}.

\subsection{Creation of a high-quality historical meteorological data for the southern south-America region}\label{sec:dataset}

\noindent\textbf{National weather stations.} The main dataset used in our project comprises data from~$137$ weather stations in southern South America. 
This dataset includes daily maximum and minimum temperatures, as well as daily accumulated precipitation from the period $1977$ to $2018$. 
This period was chosen in consultation with the institutions that provided the data, intending to obtain the largest amount of quality data.
Table~\ref{MeteoStations} provides detailed information about the number of stations for each country and the institutions from which we obtained the data. A first curated version of this dataset was used to forecast the average temperatures in~\cite{Kiedanski2025}.
%
%

\begin{figure}[h]
    \centering
    \includegraphics[width=\textwidth]{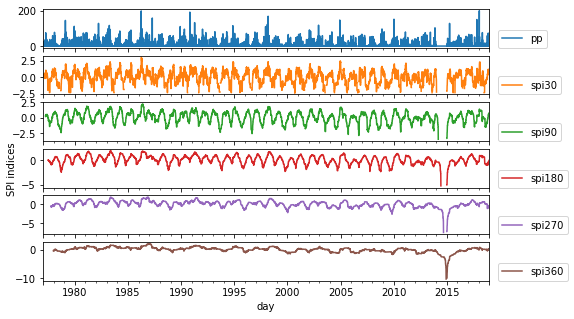}
    \caption{Cumulated precipitations and several SPI indices in a weather station located at latitude -60.45 and longitude -26.87 (ID: NH0415). The figure shows the typical frequency of SPI indices, where the peaks are not synchronized. It has a significant period of missing data in 2014, which is a common issue when dealing with historical weather data.}
    \label{spi_NH0415}
\end{figure}

\begin{figure}[h]
    \centering
        \includegraphics[width=0.475\columnwidth]{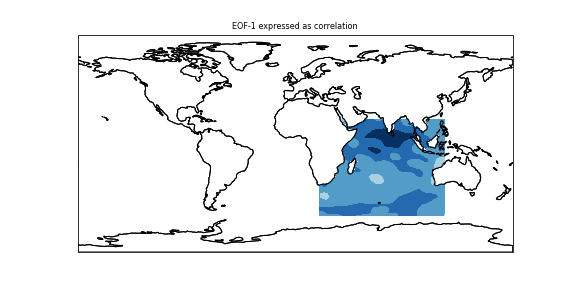}
        \includegraphics[width=0.475\columnwidth]{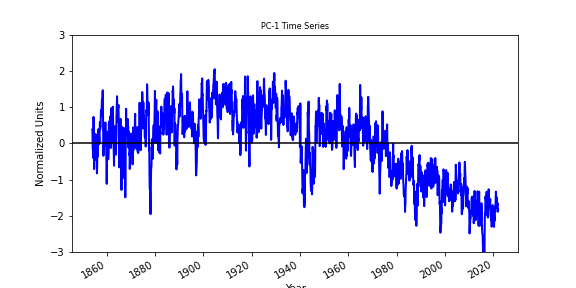}
        \includegraphics[width=0.475\columnwidth]{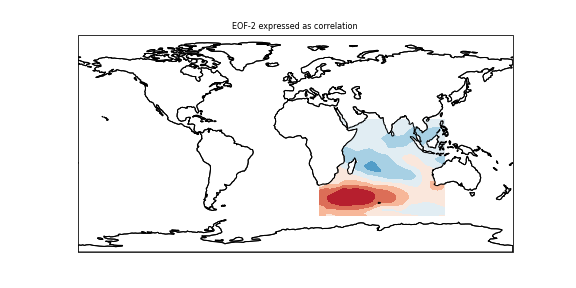}
        \includegraphics[width=0.475\columnwidth]{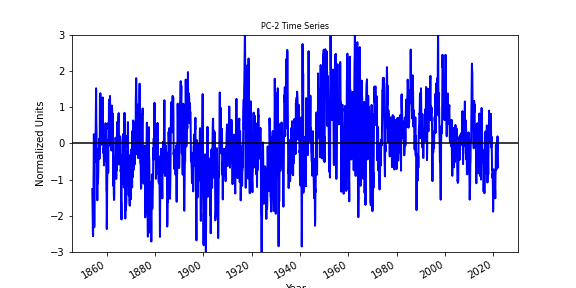}
        \includegraphics[width=0.475\columnwidth]{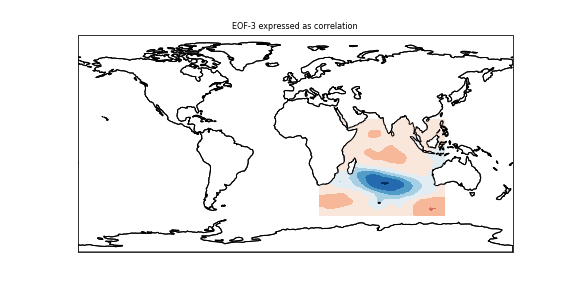}
        \includegraphics[width=0.475\columnwidth]{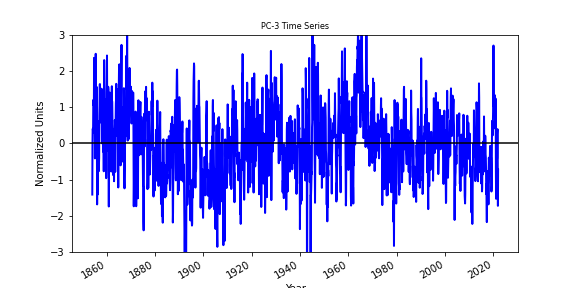}
        \includegraphics[width=0.475\columnwidth]{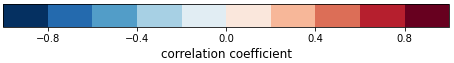}
    \caption{First three EOFs for the Pacific Ocean SST data and its associated time series.}
    \label{fig:eofs:pac}
\end{figure}

\begin{figure}[h]
    \centering
        \includegraphics[width=0.475\columnwidth]{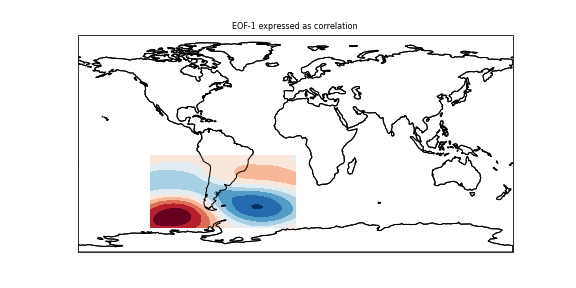}
        \includegraphics[width=0.475\columnwidth]{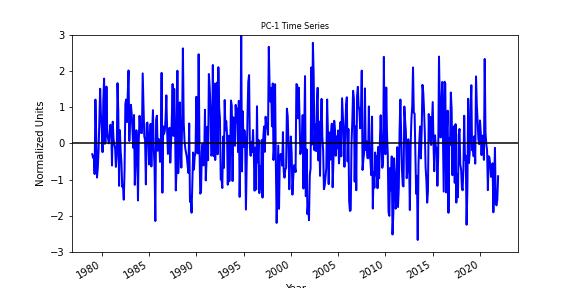}
        \includegraphics[width=0.475\columnwidth]{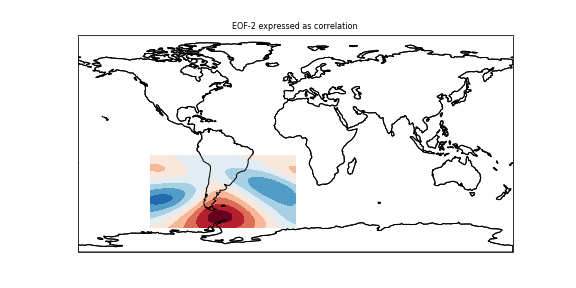}
        \includegraphics[width=0.475\columnwidth]{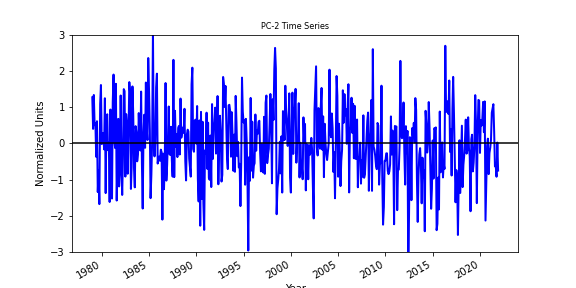}
        \includegraphics[width=0.475\columnwidth]{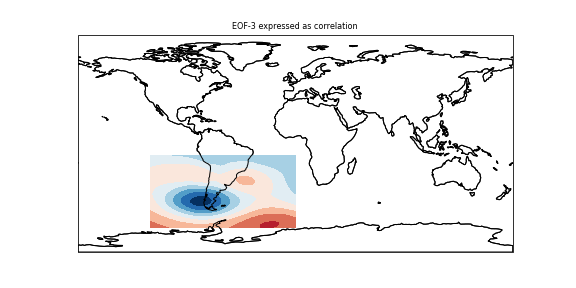}
        \includegraphics[width=0.475\columnwidth]{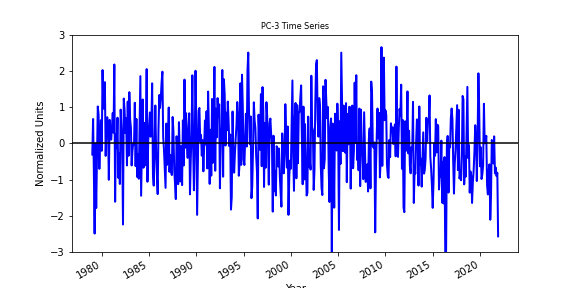}
        \includegraphics[width=0.475\columnwidth]{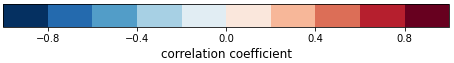}
    \caption{First three EOFs for the region spanning longitudes 240° to 360°E and latitudes -70° to -10°S, derived from HGT500 data and its associated time series.}
    \label{fig:eofs:hgt500}
\end{figure}

\noindent\textbf{Selected features.}  In this study, we considered a wide range of weather variables and trends:
\begin{itemize}
\item {Daily features:} The considered variables are: maximum temperature~$t_{max}$, minimum temperature~$t_{min}$, mean temperature~$\overline{y}$, average mean temperature over the last 90 days~$\overline{y_\text{90}}$ in Celsius, and accumulated precipitations~$p$ in mm.
\item {Static weather station features:} Per each station, we consider the spatial information of \textit{longitude}, \textit{latitude}, and \textit{altitude} (in meters).
\item {Cyclical features:} To capture the cyclical nature of seasonal patterns, we transform the day of the year into two orthogonal sinusoidal real values~$year_{sin}$ and $year_{cos}$.
\item {Standardized Precipitation Indicators (SPI):} Using daily precipitation data, we compute several SPI indices, a widely used variable to quantify information on drought conditions, for the following $30$, $90$, $180$, $270$, and $360$ days timescales. The resulting indices are denoted as $spi30$, $spi90$, $spi180$, $spi270$ and $spi360$, respectively. 
We present an example of these variables in Figure~\ref{spi_NH0415}.
\item  {Sea Surface Temperature (SST):} We obtained the SST data with a grid granularity of $90 \times 180$ points (lat, lon) from the Physical Sciences Laboratory, National Oceanic and Atmospheric Administration (NOAA)~\cite{NOAA_ERSST_V5}.
The SST data was used to compute a measure of anomalous monthly temperature at each grid point, as the difference between the current temperature and the historical average temperature (from 1981 to 2010) at this grid point.
As an example, the anomaly of January 2018 is computed as the difference of the current mean temperature in January 2018 with the average mean temperature for January during the period from 1981 to 2010. Moreover, we separate the SST data by ocean basins, defined as: Pacific Ocean (Longitudes from 125 to 290°E, latitudes from -70 to 60°N), Atlantic Ocean (Longitudes from 290 to 340°E, latitudes from -70 to 70°N), and Indian Ocean (Longitudes from 20 to 125°E, latitudes from -60 to 20°N).
We orthogonally decompose these three time series of anomalies over a map grid, we compute the main spatial variability patterns using the Empirical Orthogonal Function (EOF) analysis~\cite{eof_ZHANG2015161}.
We compute the first~$20$ EOFs for each time series and select the minimum number of EOFs required to explain $50\%$ of the variance, using these as features. We selected 7, 6 and 3 EOFs for the Pacific, Atlantic, and Indian Ocean basins, respectively.
These time series are used as time-dependent features. Therefore, in the Pacific Ocean case, we have seven new features. These features summarize important climate predictors~\cite{Kim2020}, such as El Niño-Southern Oscillation (ENSO) in the first EOF, and the  Pacific Decadal Oscillation (PDO) in the second EOF.
%
Following the same example of the Pacific Ocean, Figure~\ref{fig:eofs:pac} shows the first three EOFs and its associated time series. 
%

\item Geopotential Height 500 hPa (HGT500):
We obtained from European Center for Medium-Range Weather Forecasts (ECMWF) the daily values at 500 hPa between the spanning longitudes from 240° to 360°E and latitudes from -70° to -10°S.
Then, we proceed to generate time series anomalies using a similar approach to that used for the SST data. We first compute the daily anomaly at each grid point, as the difference between the current and the historical average (from 1981 to 2010) at this grid point. Then, with this time series of anomalies over a map grid, we compute the first 20 EOFs, and we keep as features the first three EOFs that explain $50\%$ of the variance. See the results in Figure~\ref{fig:eofs:hgt500}.
\end{itemize}
\begin{table}[h!]
\caption{Information about the weather stations.}
\centering
\scalebox{1}{
\begin{tabular}{llcc}
\hline\hline
\multicolumn{1}{c}{Country} & \multicolumn{1}{c}{Institution} & \multicolumn{1}{c}{Stations} & \multicolumn{1}{c}{Source} \\
\hline
Argentina &  Servicio Meteorológico Nacional & $71$ & \cite{smn_argentina} \\
Brasil & Instituto Nacional de Meteorología & $10$ & \cite{smn_brasil} \\
Chile & Centro de Investigaciones sobre Clima y Resiliencia & $39$ & \cite{smn_chile}  \\
Paraguay & Dirección de Meteorología e Hidrología & 7 &\cite{smn_paraguay}  \\
Uruguay & Instituto Uruguayo de Meteorología & 10 &\cite{smn_uruguay} \\
\hline
\hline
\end{tabular}
}
\label{MeteoStations}
\end{table}
\begin{figure}[h]
\centering \includegraphics[width=1.05\textwidth]{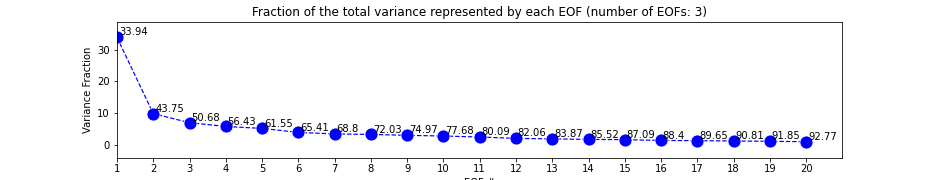}
\caption{Fraction of total variance represented by the EOF in the Pacific ocean basing.}
    \label{fig:neofs}    
\end{figure}
\subsection{Data handling and learning process}

The training data includes the first $40$ years, from $1977$ to $2016$; the year $2017$ is used for testing and model evaluation. 
In general, the dataset has approximately~$4.8\%$ of missing temperature data from the regional weather stations. However, the missing data appears differently depending on the geographical point and date. 
%
Weather stations with a high level of missing data were discarded. We discarded 69 of the 206 analyzed stations, meaning~$33\%$ of the original weather stations were excluded due to unreliable information.
%
In addition, the data collected during the year $2018$ has~$19.4\%$ of missing temperature data, therefore this year is discarded in the statistical learning analysis for being different to the data used for training. 

Considering that the testing dataset is quite large ($12$ months), we adopted the following approach: We train $10$ models. At the start of each testing month, we train a new model with the historical information until the date, and evaluate the performance in the next three months (approx. 90 days). 

\subsection{Setting of the AutoGluonTS framework}
\label{AutoGluonTS}
\noindent\textbf{Studied forecasting models.}
The AutoGluonTS tool automatically trains multiple state-of-the-art forecasting  models on a given multidimensional time series~\cite{Erickson2023}.
An advantage of this tool is its AutoML capabilities, which streamline the entire forecasting process, 
eliminating the need for manual tasks such as data cleaning, hyperparameter optimization, feature and model selection.
AutoGluonTS has three categories of forecasting techniques: local, global, and ensemble models. 
Local methods are used as baseline, the approach fits  separate models to each time series, capturing patterns as trend and seasonality.
Global models learn from the entire training multidimensional dataset to predict multiple time series. 
Ensemble models combine outputs from multiple predictors, including local and global models, in a multi-layer approach where the combined outputs are fed back into the models. Read~\cite{Erickson2023} for details. 
This iterative process refines the predictions with the number of iterations. We incorporate the following local methods as forecasting tools in our experimental analysis of AutoGluonTS.
\begin{itemize}
\item Na\"ive and SeasonalNa\"ive: The Na\"ive technique, often used as a baseline, refers to the extreme solution of predicting the next value of a time series by copying the present value.
The SeasonalNa\"ive approach predicts the value one time step ahead using the observed value of last year at the time step.
\item Error, Trends and Seasonality (ETS): The ETS models extend  Exponential Smoothing techniques by including error, trends and seasonality explicitly in the model. They are statistical-based methods that compute weighted averages of the previously observed values, where the weights exponentially decrease with the time lags~\cite{Hyndman21}. 
In addition, AutogluonTS includes AutoETS, which an ETS forecasting model with partially automatic parametrization.
\item {Theta:} This is a statistical method powerful to capture both central tendencies and trends, as well as to separate short-term and long-term segments of a time series~\cite{theta00}. 
\item Autoregressive Integrated Moving Average (ARIMA): This is a classic statistical method used for time series forecasting. It includes a differencing step to transform a non-stationary time series into a stationary one.
AutogluonTS also has variations of the original ARIMA: Seasonal ARIMA (SARIMA) and SARIMA with Exogenous Variables (SARIMAX). These two techniques eliminate seasonal effects and integrate exogenous variables.
\end{itemize}
In addition, we also include the following two global models in the experimental evaluation of AutoGluonTS:
\begin{itemize}
\item AutoGluonTabular: 
The forecasting task is converted into a tabular problem using temporal features.
Considering this structured data the model solves the forecasting task by applying gradient-boosted tree algorithms such as XGBoost, CatBoost and LightGBM.
The structured data is constructed using the features:
\begin{itemize}
\item Lag features: Observed time series values based on the frequency of the training data. 
\item Time features: It refers to temporal variables, for instance, day of the year based on the timestamp of the measurement. \item Lagged known and past covariates.
\item Static features:  For instance, longitude, latitude, altitude, and so on. 
\end{itemize}
\item {Deep Auto Regressive (DeepAR):} This technique, based in a recurrent NNs, is an autoregressive forecasting model that produces probabilistic forecasts over a large number of related time series in a training dataset~\cite{deepar_SALINAS20201181}.
Instead of single-valued forecasts, the model output is a probabilistic forecasting model.
\end{itemize}
Furthermore, we evaluate a weighted ensemble where the predictions are weighted according to the accuracy of single predictions.

\noindent\textbf{Hyperparametric setting.}
We investigate a configuration of the AutoGluonTS framework, where only a few selected parameters are given by us, which are~\cite{Erickson2023}: (i) The hyperparameter optimization for local statistical models \textit{preset} is set to ``high\_quality''. (ii) We limit the maximum training time to~$4$ hours  and discard the results if the framework does not finish within this time limit. (iii) We set the forecasting horizon to $90$ time units (day-scale). (iv) We set the performance metric used for ranking the models inside the valiation procedure of AutoGluonTS as the Symmetric Mean Absolute Percentage Error (sMAPE).

\noindent\textbf{Computational details.} We run the AutoGluonTS V1.1 framework~\cite{autogluon2023} in a high performance server with the following main specifications: Intel(R) Xeon(R) Silver 4210 CPU @ 2.20GHz, with 40 CPU cores and  96GB RAM.
%
\section{Experimental results}\label{sec:results}
\begin{figure}[htbp]
\centering
\includegraphics[width=0.45\textwidth]{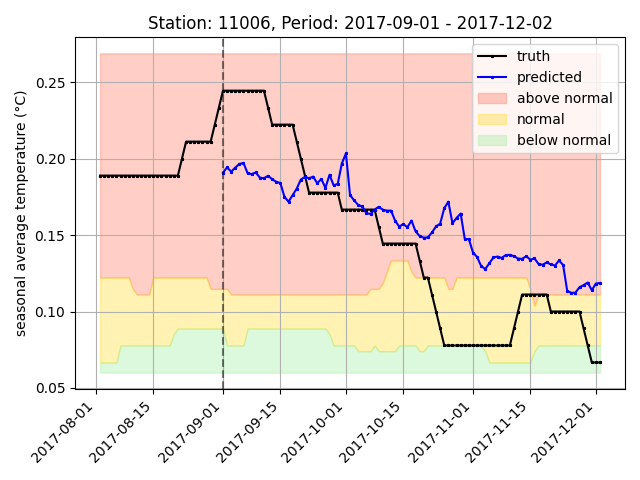}
\includegraphics[width=0.45\textwidth]{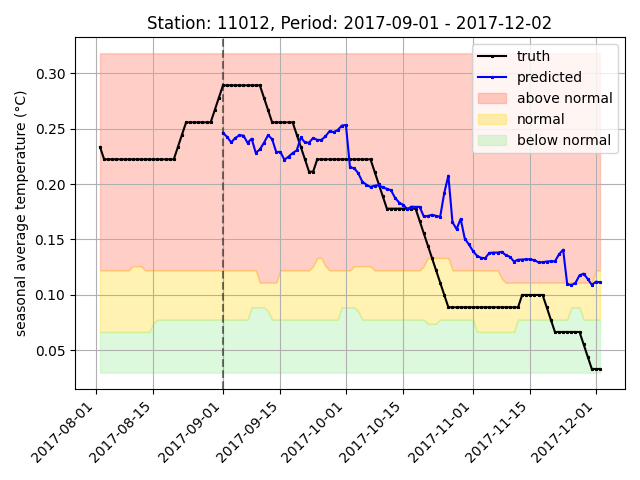}
\includegraphics[width=0.45\textwidth]{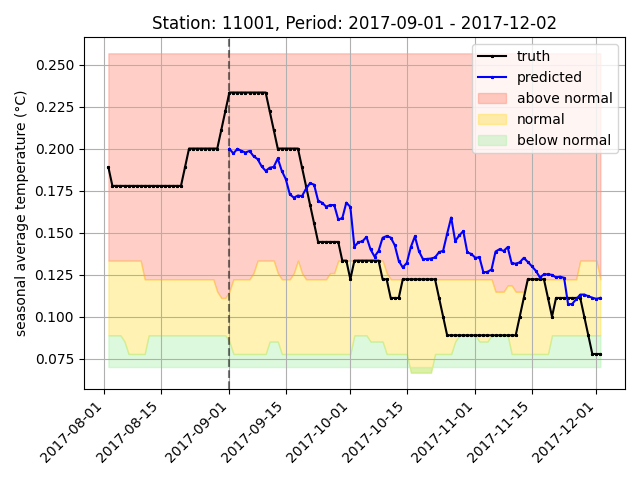}
         \includegraphics[width=0.45\textwidth]{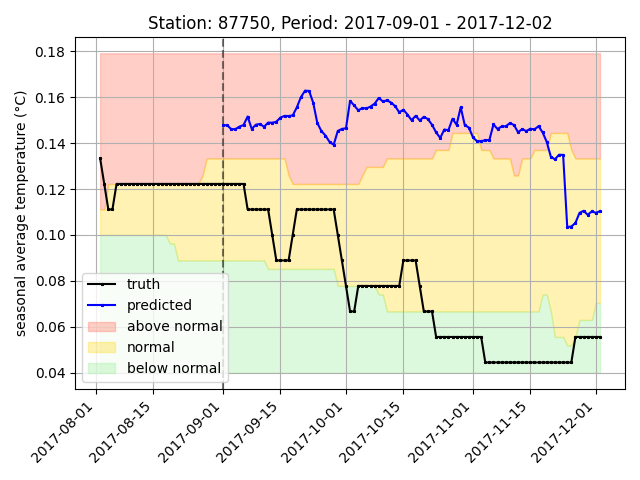}
    \caption{\label{Midlatitude} Example of predictions for four weather stations located in midlatitude climates.}
\end{figure}
\begin{figure}[htbp]
    \centering
    \includegraphics[width=0.45\textwidth]{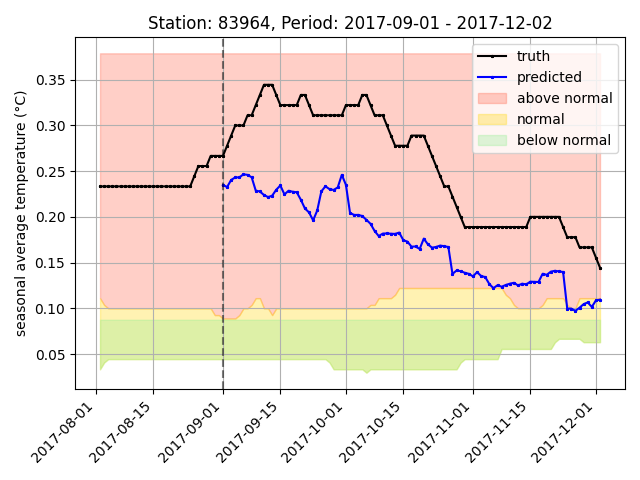}
    \includegraphics[width=0.45\textwidth]{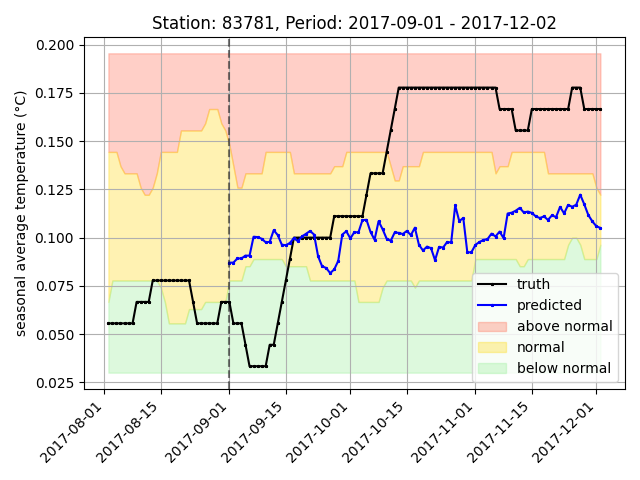}
    \hfill
    \centering
    \includegraphics[width=0.45\textwidth]{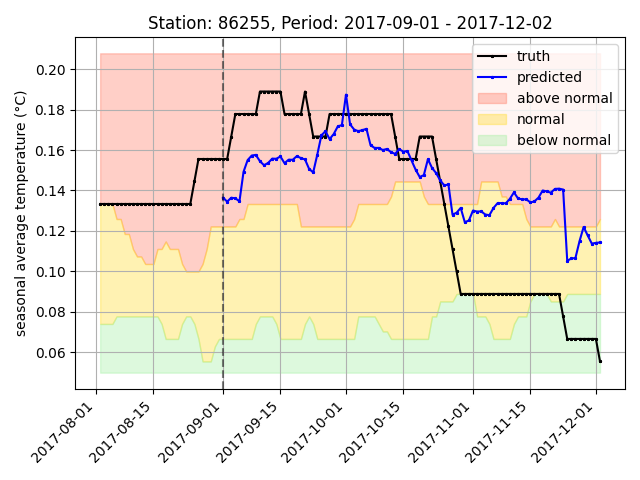}
    \caption{\label{Subtropical}Example of predictions for three weather stations located in humid subtropical climates.}
\end{figure}
\begin{figure}[htbp]
    \includegraphics[width=0.45\textwidth]{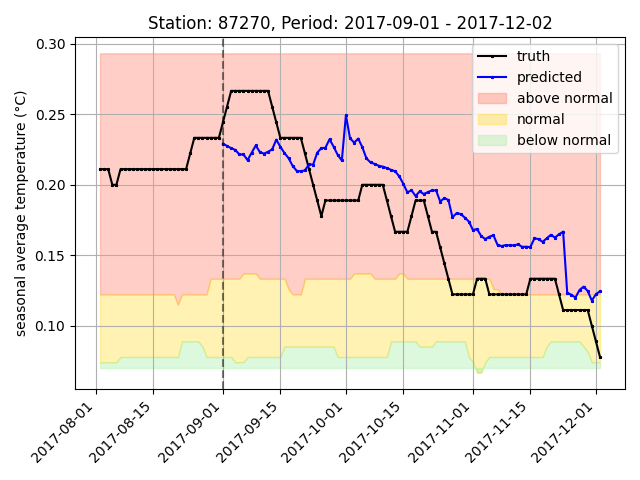}
    \includegraphics[width=0.45\textwidth]{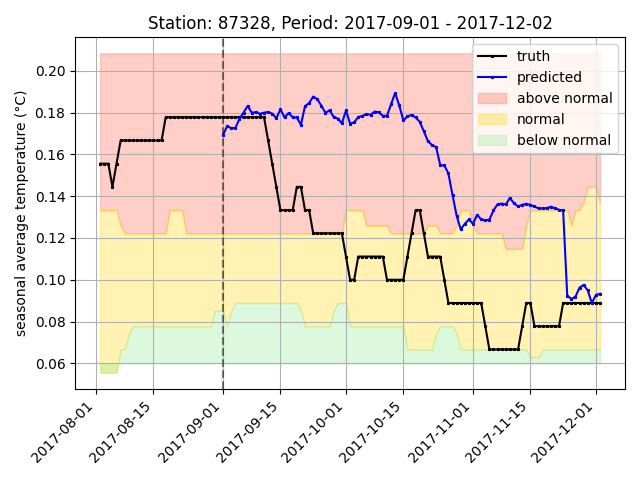}
    \hfill
    \centering
    \includegraphics[width=0.45\textwidth]{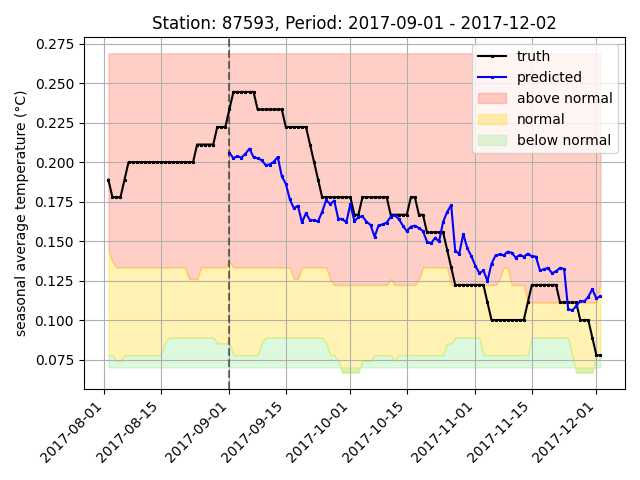}
    
    \caption{\label{SemiArid}Example of predictions for three weather stations located in different climate conditions.}
\end{figure}
\begin{figure}[htbp]
    \centering
\includegraphics[width=0.6\textwidth]{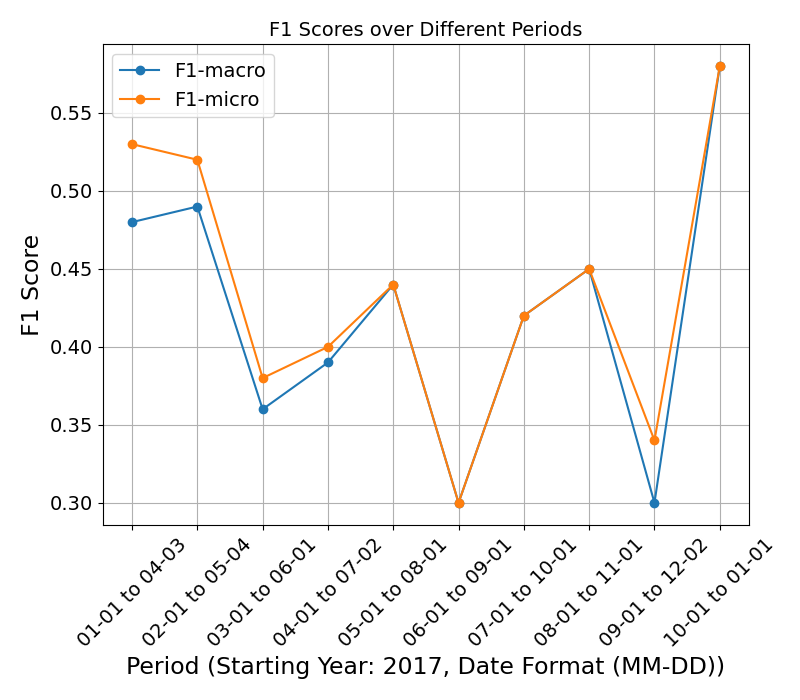}
    \caption{Global F1-macro and F1-micro scores for different periods in 2017.}
    \label{GlobalResultsF1}
\end{figure}
\begin{figure}[htbp]
    \centering
\includegraphics[width=0.6\textwidth]{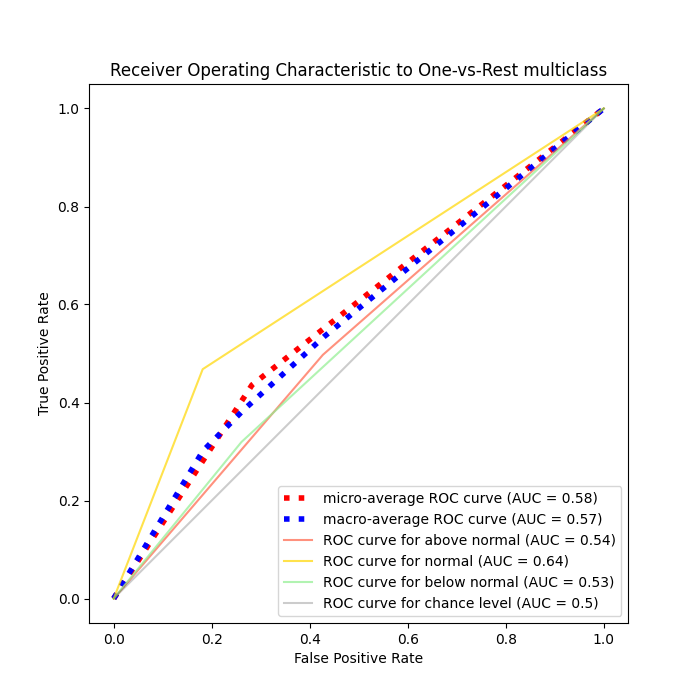}
\caption{ROC to One-vs-Rest multi-class classification and the AUC scores.}
    \label{GlobalResultsROC}
\end{figure}
After initially exploring different methods, including XGboost with intensive features engineering to be able to integrate spatial data, WaveNet, a combination of evolutionary computation and recurrent networks~\cite{Basterrech2023}, tools belonging to the Graph Neural Network family, a time series method including adaptation and distribution shift detection~\cite{Basterrech2022SMC}; we finally selected the AutoGluon platform. 
The initial exploration showed significant difference between the AutoGluon and the above mentioned, therefore, we focus the study exploring the potential of the last one. In addition, we successfully used AutoGluonTS in a preliminary work for forecasting average seasonal temperatures~\cite{Kiedanski2025}.

Figures~\ref{Midlatitude},~\ref{Subtropical}, and~\ref{SemiArid} present examples of performance of TX90w90 predictions for the first nine months in three different regions (to have information about the stations, email the authors)
These figures have a background in red, yellow and green colors, representing the tercile for the classes: {above normal}, {normal} and {below normal}.
The black line is the actual value, $\overline{y}_{w90,t+90}$ (percentage of warm days in the following $90$ days) and the blue line is the model prediction $\widehat{y}_{w90,t+90}$ for each $t$ in the following three months.
Figure~\ref{Midlatitude} has four examples corresponding to four stations located in different mid-latitude climates.
Figure~\ref{Subtropical} has three graphics corresponding to three stations located in humid subtropical regions.
And Figure~\ref{SemiArid} has three graphics with results related to three other stations located in different climates.
As expected, a common pattern is that the model gradually looses quality when it is evaluated further away from the training data. 

Figure~\ref{GlobalResultsF1} presents the evolution of micro and macro F1-scores across different periods in the year 2017.
Furthermore, Figure~\ref{GlobalResultsROC} shows the results of the analysis of the TP rate versus the FP rate. The graphic shows the ROC curve for each of the classes.
The results using the AUC scores are: for above normal (0.54), normal (0.64) and below normal (0.53).
It is challenging to make direct comparisons, but notice that regions and datasets are totally different. So, just observing our results, you see that they appear competitive with the state-of-the-art models used in industry~\cite{NMME,SMN-CPT,wwwTX902025}.
The results that most closely resemble the conditions of our work are those presented in~\cite{wwwTX902025}. 
This work uses a quarterly forecast (similar to our 90-day forecast), but while we do so with a rolling window for each day, this benchmark work uses a three-month forecast. Therefore, they have 90 times fewer evaluation dates than ours.
This reference also provides forecasts only for Argentina, 71 of the 137 stations in our study. This makes the comparison even more difficult, since in our work we verified that predictability depends largely on the geography and latitude and longitude of the meteorological station.
Quality metrics include precision (not F1 score nor AUC), with a typical accuracy of 0.22-0.28 for the TX90 forecast\footnote{See details in: http://pronosticosextremos.at.fcen.uba.ar/verificación.html, accessed: 2025-03-
28.}.
This quarterly climate forecast is based on the analysis of numerical experimental predictions from the main global climate simulation models and Argentinian statistical models, combined with expert analysis of the evolution of oceanic and atmospheric conditions. Then, this forecast is based on a consolidated manual consensus from these various sources and is, therefore, impossible to reproduce automatically\footnote{See details in: https://www.smn.gob.ar/pronostico-trimestral, accessed: 2025-03-
28.}.
As future work, we include the possibility of conducting a joint project with the Argentine National Meteorological System to compare the performance of both models.
%
%

\section{Conclusions and future work}\label{sec:conclusions}
In this paper, we address the problem of predicting future values of maximum daily temperatures over a medium to long term horizon. We created a historical dataset with spatiotemporal information covering~$40$ years of weather data from over~$130$ weather stations in the southern region of South America. In addition, we incorporate exogenous variables from ocean basins.
The results obtained using the AutoGluonTS framework are similar to those achieved with state-of-the-art, large platforms implementing classical statistical and physical methodologies. However, we used a standard computation server, unlike the substantial resources employed by major research centers providing this type of information today. This shows that the machine learning approach is a promising choice for this class of problem.

We plan to continue our research in three parallel directions: 
\begin{itemize}
    \item One is to provide a large comparison in performance between our model and state-of-the-art models. As mentioned above, this task is challenging due to the significant differences in geographic and temporal evaluation points with the reference forecast provided.
    \item Another direction is related to improving the computational performance and the data quality, while keeping the goal of maintaining a low cost of execution for the models. 
    The enrichment can include other exogenous variables, especially those related to the oceans, and possibly statistical information. In addition, we plan to update the version of AutoGluonTS, which certainly deserves exploration.
    \item The last direction focuses on causality and explainability. There are regions and specific weather stations with historical data that are harder to model than others. We plan to conduct a deeper study to explain the reasons behind the differences in the results among the time series.
\end{itemize}

%
%
%
%
%
\section*{Acknowledgments}
This work was supported by the ``ClimateDL'' project belonging to the Climat AmSud program, with program code 22-CLIMAT-02. S. Collazo is funded by the European Union’s Horizon 2020 research and innovation program under the Marie Skłodowska-Curie grant agreement No 847635 (UNA4CAREER) through the SAFETE project (code 4230420).

%
%

%
\bibliographystyle{unsrt}
\bibliography{bibliography}
\end{document}